\documentclass[10pt, conference, compsocconf]{IEEEtran}
\usepackage{cite}
\usepackage{amsmath,amssymb,amsfonts}
\ifCLASSINFOpdf
\else
\fi
\usepackage{algorithm}
\usepackage{algorithmic}
\usepackage{multirow}
\usepackage{graphicx}
\usepackage{textcomp}
\begin{document}
\title{ Parsimonious HMMs for Offline Handwritten Chinese Text Recognition }


\author{\IEEEauthorblockN{Wenchao Wang, Jun Du and Zi-Rui Wang}
\IEEEauthorblockA{University of Science and Technology of China\\
Hefei, Anhui, P. R. China\\
Email: wangwenc@mail.ustc.edu.cn, jundu@ustc.edu.cn, cs211@mail.ustc.edu.cn}}


%Telephone: (800) 555--1212, Fax: (888) 555--1212}


\maketitle

\begin{abstract}
Recently, hidden Markov models (HMMs) have achieved promising results for offline handwritten Chinese text recognition. However, due to the large vocabulary of Chinese characters with each modeled by a uniform and fixed number of hidden states, a high demand of memory and computation is required. In this study, to address this issue, we present parsimonious HMMs via the state tying which can fully utilize the similarities among different Chinese characters. Two-step algorithm with the data-driven question-set is adopted to generate the tied-state pool using the likelihood measure. The proposed parsimonious HMMs with both Gaussian mixture models (GMMs) and deep neural networks (DNNs) as the emission distributions not only lead to a compact model but also improve the recognition accuracy via the data sharing for the tied states and the confusion decreasing among state classes. Tested on ICDAR-2013 competition database, in the best configured case, the new parsimonious DNN-HMM can yield a relative character error rate (CER) reduction of 6.2\%, 25\% reduction of model size and 60\% reduction of decoding time over the conventional DNN-HMM. In the compact setting case of average 1-state HMM, our parsimonious DNN-HMM significantly outperforms the conventional DNN-HMM with a relative CER reduction of 35.5\%.

\end{abstract}


% \fi
\IEEEpeerreviewmaketitle

\section{Introduction}
Offline handwritten Chinese text recognition (HCTR) is a challenge topic due to large vocabulary and unrestrained writing styles~\cite{yin2013icdar}. Most existing techniques can be classified into two categories: oversegmentation-based and segmentation-free approaches. Oversegmentation-based approaches often need to explicitly segment text line into a sequence of primitive image patches and then merge them to form a candidate lattice~\cite{fu2006novel,li2010bayesian,wang2012handwritten,wu2017improving}. In contrast to the oversegmentation-based approaches, segmentation-free approaches do not require the explicit segmentation for text line. \cite{su2009off} adopted the Gaussian mixture model based hidden Markov model (GMM-HMM) for the text line modeling. With the success of deep learning~\cite{lecun2015deep}, deep neural networks (DNNs) have been widely applied for HCTR. Recently, \cite{messina2015segmentation} successfully used multidimensional long-short term memory recurrent neural network (MDLSTM-RNN) with connectionist temporal classification (CTC)~\cite{graves2006connectionist} for HCTR. More recently, \cite{du2016deep,wang2016writer,wang2017deep} proposed hybrid neural network based HMMs (NN-HMMs) for HCTR, which achieved the best performance on the ICDAR-2013 competition database \cite{yin2013icdar} among existing segmentation-free approaches.

\begin{figure}
	\centering
	\setlength{\abovecaptionskip}{0pt}
	\setlength{\belowcaptionskip}{0pt}
	\includegraphics[width=0.5\textwidth]{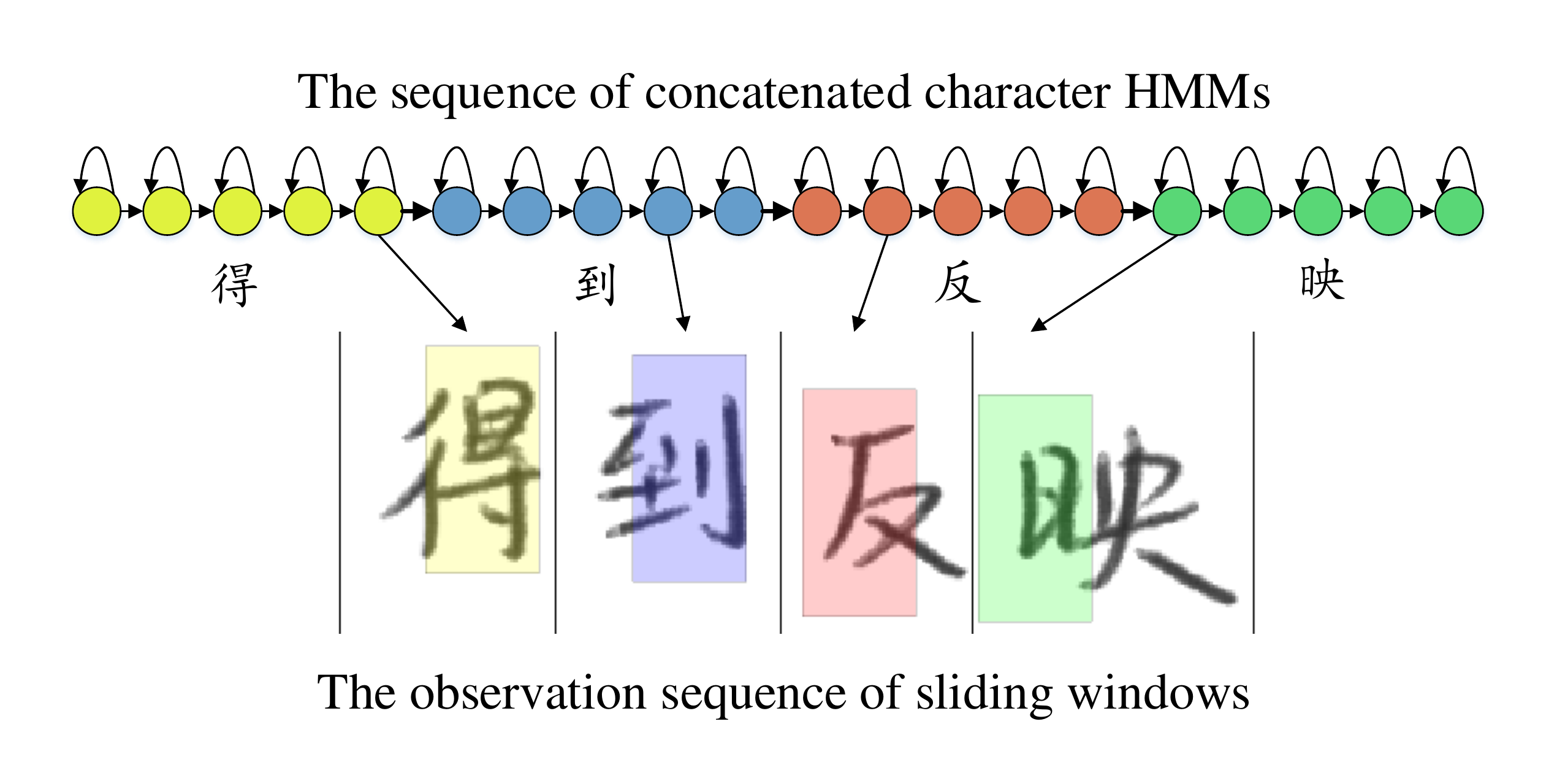}
	\caption{HMM-based handwritten Chinese text line modeling.}
	\label{fig:one}
\end{figure}

\begin{figure}
	\centering
	\setlength{\abovecaptionskip}{0pt}
	\setlength{\belowcaptionskip}{0pt}
	\includegraphics[width=0.53\textwidth]{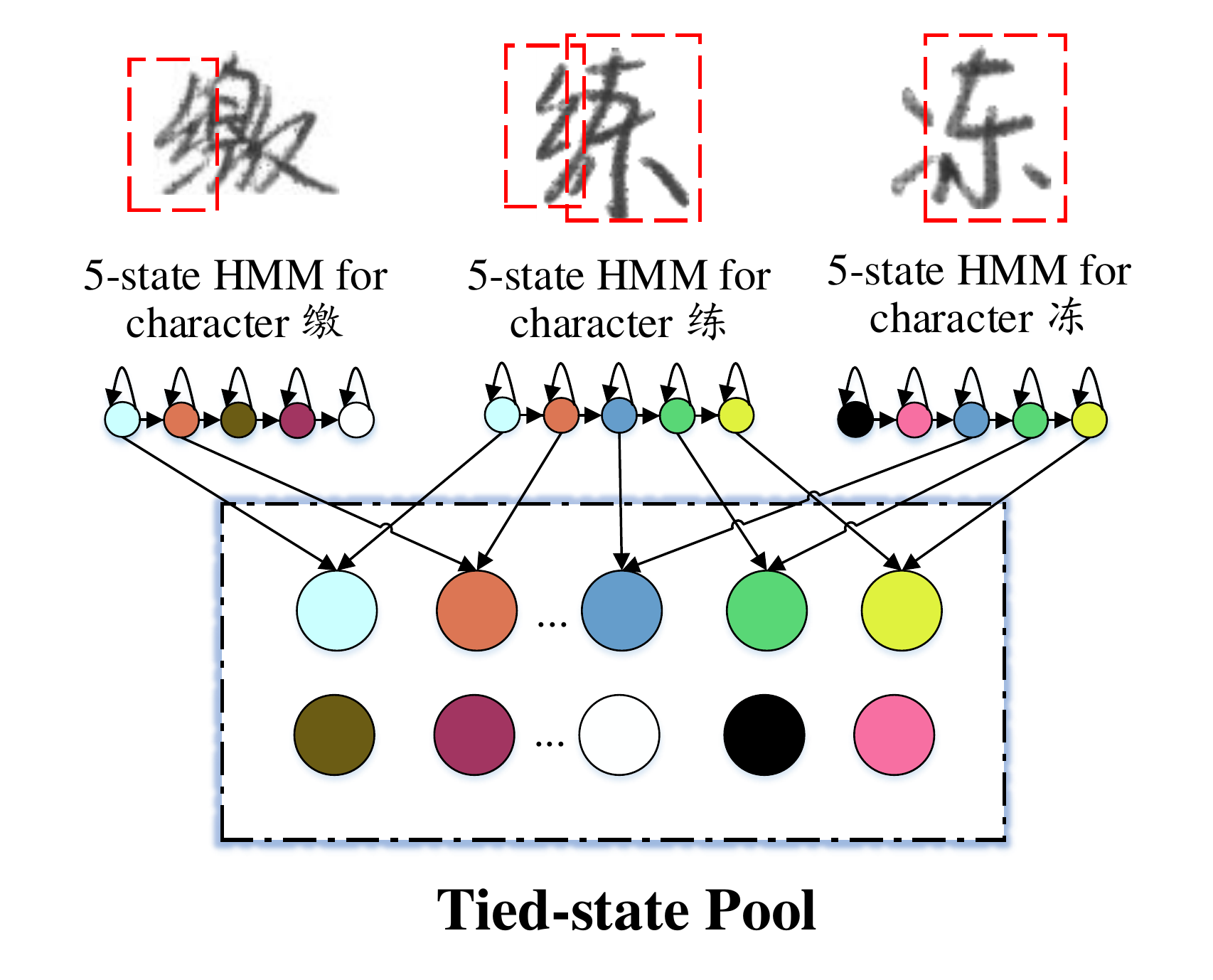}
	\caption{ Illustration of state tying.}
	\label{fig:two}
\end{figure}

The success of NN-HMMs \cite{du2016deep,wang2017deep} is attributed to two aspects. First, the DNN or convolutional neural network (CNN) \cite{lecun1998gradient} is powerful in modeling the emission distributions just like in MDLSTM-RNN \cite{messina2015segmentation}. Second, the left-to-right HMM \cite{rabiner1989tutorial} with a set of hidden states is adopted to represent each character class, illustrated in Fig.~\ref{fig:one}. %, yielding a higher modeling resolution than CTC \cite{graves2006connectionist} quite similar to 1-state HMM. This is why MDLSTM-RNN \cite{messina2015segmentation} using a more powerful network architecture than DNN still underperforms DNN-HMM as reported in \cite{du2016deep}.
Accordingly, to model the text line as a observation sequence of frames implemented by sliding windows, the character HMMs are concatenated as shown in Fig.~\ref{fig:one}. However, there is one main problem in the conventional HMM-based HCTR, where each character is modeled with a uniform and fixed number of hidden states, e.g., 5 states in Fig.~\ref{fig:one}. Due to the large vocabulary of Chinese characters, this setting requires a high demand of memory and computation. Moreover, the uniform setting of state number is unreasonable as the similarity among different characters and the diversity of appearances are not well considered. Chinese characters, which are mainly logographic and consisting of basic radicals, constitute the oldest continuously used system of writing in the world which is different from the purely sound-based writing systems~\cite{jurafsky2014speech} such as Greek, Hebrew, etc. For example in Fig.~\ref{fig:two}, the regions in red dashed boxes of the left and middle handwritten Chinese characters are quite similar as they belong to the same radical.

In this study, to address the above-mentioned problem in conventional DNN-HMM approach, we present parsimonious DNN-HMMs via the state tying which can fully utilize the similarities among different Chinese characters. We adopt two-step algorithm with the data-driven question-set to generate the tied-state pool using the likelihood measure, which is inspired by the similar idea in speech recognition area \cite{bahl1991decision,young1994tree,dan2012}. The proposed parsimonious DNN-HMMs not only lead to a compact model but also improve the recognition accuracy via the data sharing for the tied states and the confusion decreasing among state classes. Tested on ICDAR-2013 competition database, in the best configured case, the new parsimonious DNN-HMM can yield a relative character error rate (CER) reduction of 6.2\%, 25\% reduction of model size and 60\% reduction of decoding time over the conventional DNN-HMM. In the compact setting case of average 1-state HMM, our parsimonious DNN-HMM significantly outperforms the conventional DNN-HMM with a relative CER reduction of 35.5\%.

\section{Overview of Parsimonious DNN-HMM}

The proposed framework aims to search the optimal character sequence $\hat{\mathbf{C}}$ for a given extracted feature sequence $\mathbf{X}=\{\mathbf{x}_0,\mathbf{x}_1,...,\mathbf{x}_T\}$ of a text line, which can be formulated according to the Bayesian decision theory as follows:
\begin{eqnarray}
\hat{\mathbf{C}}=\arg\max\limits_{\mathbf{C}}p(\mathbf{C}|\mathbf{X})=\arg\max\limits_{\mathbf{C}}p(\mathbf{X}|\mathbf{C})P(\mathbf{C})
\end{eqnarray}
where $p(\mathbf{X}|\mathbf{C})$ is the conditional probability of $\mathbf{ X}$ given $\mathbf{C}$ which is named as the character model. Meanwhile $P(\mathbf{C})$ is the prior probability of $\mathbf{C}$ which is named as the language model. As one implementation of this Bayesian framework, we use an HMM to model one character class, accordingly a text line is modeled by a sequence of HMMs. An HMM has a set of states and each frame is supposed to be assigned to one underlying state. For each state, an emission distribution describes the statistical property of the observed frame. With HMMs, we rewrite the $p(\mathbf{X}|\mathbf{C})$ in:
\begin{eqnarray}
p(\mathbf{X}|\mathbf{C}) &=& \sum_S\left[p(\mathbf{X},S|\mathbf{C})\right]\\
~ &=& \sum_S\left[\pi(s_0)\prod\limits_{t=1}^Ta_{s_{t-1}s_t}p(\mathbf{x}_t|s_t)\right]\\
~ &=& \sum_S\left[\pi(s_0)\prod\limits_{t=1}^Ta_{s_{t-1}s_t}\frac{p(s_t|\mathbf{x}_t)p(\mathbf{x}_t)}{p(s_t)}\right]
\end{eqnarray}
$S = \{s_0,s_1,...,s_T\}$ is one underlying state sequence of $\mathbf{C}$ to represent $\mathbf{ X}$. $\pi(s_0)$ is the prior probability of the initial state $s_0$ and $a_{s_{t-1}s_t}$ is the transition probability from state $s_{t-1}$ at the $(t-1)^{\textrm{th}}$ frame to state $s_t$ at the $t^{\textrm{th}}$ frame. $p(\mathbf{x}_t|s_t)$ is the emission probability, which can be directly calculated (e.g., GMM in \cite{su2009off}) or indirectly obtained via the state posterior probability $p(s_t|\mathbf{x}_t)$ (e.g. DNN in \cite{du2016deep}).

Within this framework, the main procedure to train parsimonious DNN-HMMs are summarized in Algorithm \ref{alg:A}. In the recognition stage, after the feature extraction of the unknown handwritten text line, the final recognition results can be generated via a weighted finite-state transducer (WFST)~\cite{mohri2002weighted,allauzen2007openfst} based decoder by integrating both character model and language model. Note that the number of output layer neurons in DNN corresponds to the number of tied-states, which is controlled by state tying results. In the next section, we will elaborate the state tying algorithm.

\begin{algorithm}
	\caption{Training steps of parsimonious DNN-HMMs}
	\label{alg:A}
	\begin{algorithmic}
		\item[1] Preprocess all training text lines and extract gradient direction features~\cite{liu2007normalization} followed by PCA transformation~\cite{rencher2003methods}.
		\item[2] Train conventional GMM-HMMs with the uniform and fixed number of hidden states for all character HMMs.
		\item[3] Calculate the first-order and second-order statistics based on state-level forced-alignment based on GMM-HMMs.
		\item[4] Generate the question set based on the statistics using a top-down data-driven method.
		\item[5] \text{Two-step algorithm:} \\
		\begin{itemize}
			\item
			First-step: Build the state-tying trees based on statistics and question set using a top-down data-driven method.
			\item
			Second-step: Using a bottom-up greedy algorithm to recluster the tied-state results from fist-step, then get the final tied-state pool.
		\end{itemize}
		\item[6] Train parsimonious GMM-HMMs based on the final tied-state pool for all character HMMs.
		\item[7] Train parsimonious DNN-HMMs using state-level labels from the forced-alignment of parsimonious GMM-HMMs.
	\end{algorithmic}
\end{algorithm}

\section{Two-step state tying}
%\subsection{Illustration of state tying}

To give a better explanation of state tying, Fig.~\ref{fig:two} shows an example of three Chinese characters with the final tied-state pool. Each character in this figure is initially modeled by an HMM with 5 states. After parsimonious modeling, the first two states of left and middle characters are tied together while the last three states of middle and right characters are tied together. This is reasonable as these tied states are corresponding to the similar regions of dashed boxes.

We adopt two-step algorithm with the data-driven question-set to generate the tied-state pool. We will introduce the first-step algorithm, second-step algorithm and question-set building separately in the next subsections.

\subsection{The first-step data-driven method for state tying}

\begin{figure}
	\centering
	\setlength{\abovecaptionskip}{0pt}
	\setlength{\belowcaptionskip}{0pt}
	\includegraphics[width=0.5\textwidth]{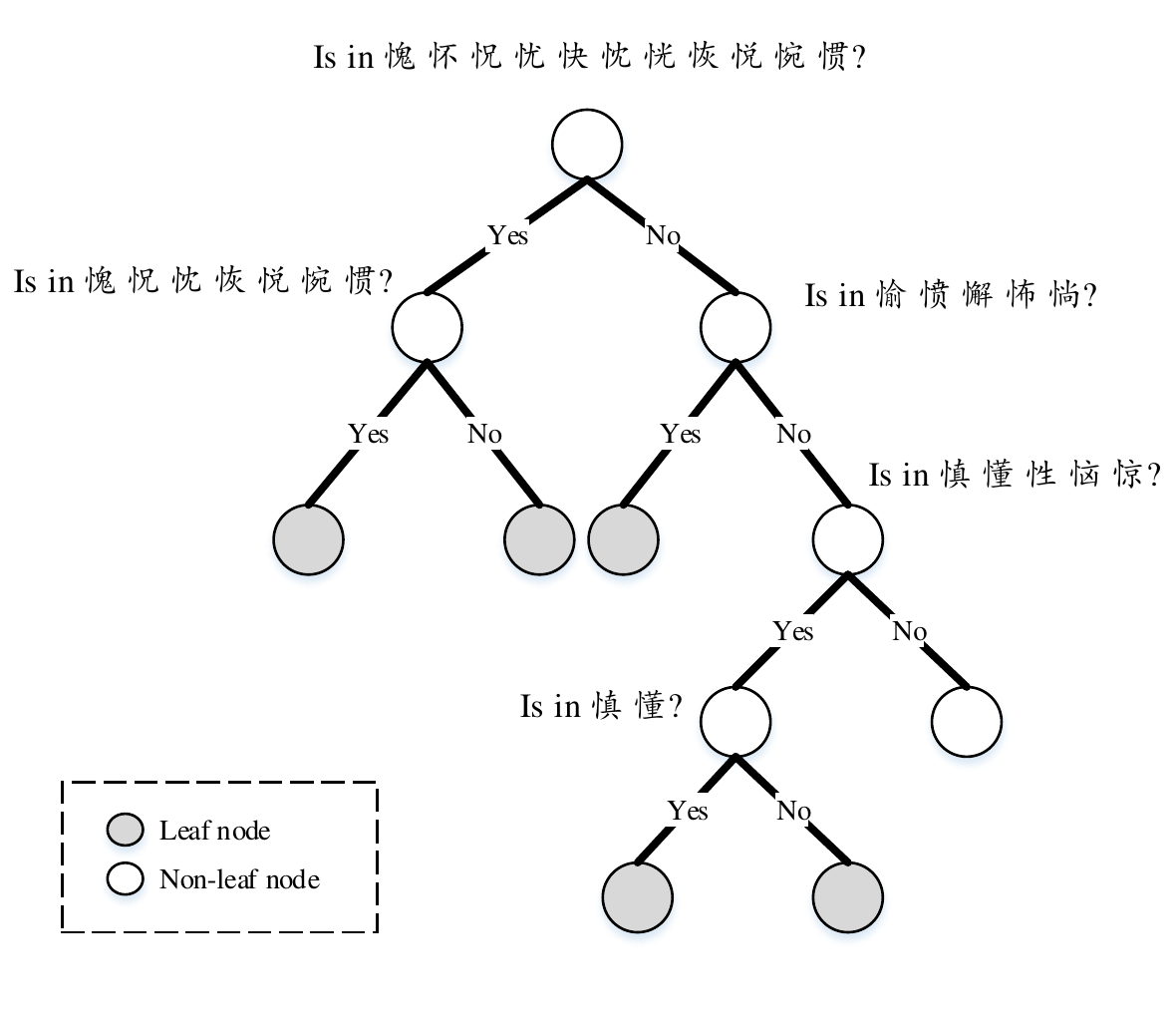}
	\caption{A tree fragment for tying the first state of HMM.}
	\label{fig:three}
\end{figure}

In the first step, the binary decision tree is adopted for state tying with each node partitioned by a question. Each question is related with a set of Chinese characters which will be described in Section III-C. One tree is constructed for each HMM state (e.g., state 1 to state 5 in Fig.~\ref{fig:two}) to cluster the corresponding states of all associated characters. Because the number of Chinese characters used in this study is 3980, the whole tree on each state is pretty large. In Fig.~\ref{fig:three}, we just show a fragment of the decision tree for tying the first state of HMM, where five clusters correspond to five leave nodes with each associated with a set of tied character classes. Similar to~\cite{young1994tree}, the basic principle is to partition states recursively to maximize the increase in expected log-likelihood. All states with the same position in HMMs are initially grouped together at the root node and the expected log-likelihood of the training data is calculated. This node is then split into two subsets based on the question which partitions the states to maximize the increase in expected log-likelihood. A maximum priority queue is maintained to save the expected log-likelihood improvements by splitting each parent node to two children nodes. Each node is then recursively partitioned until reaching the threshold of tied-state number.

\subsection{The second-step data-driven method for state tying}

\begin{figure}
	\centering
	\setlength{\abovecaptionskip}{0pt}
	\setlength{\belowcaptionskip}{0pt}
	\includegraphics[width=0.5\textwidth]{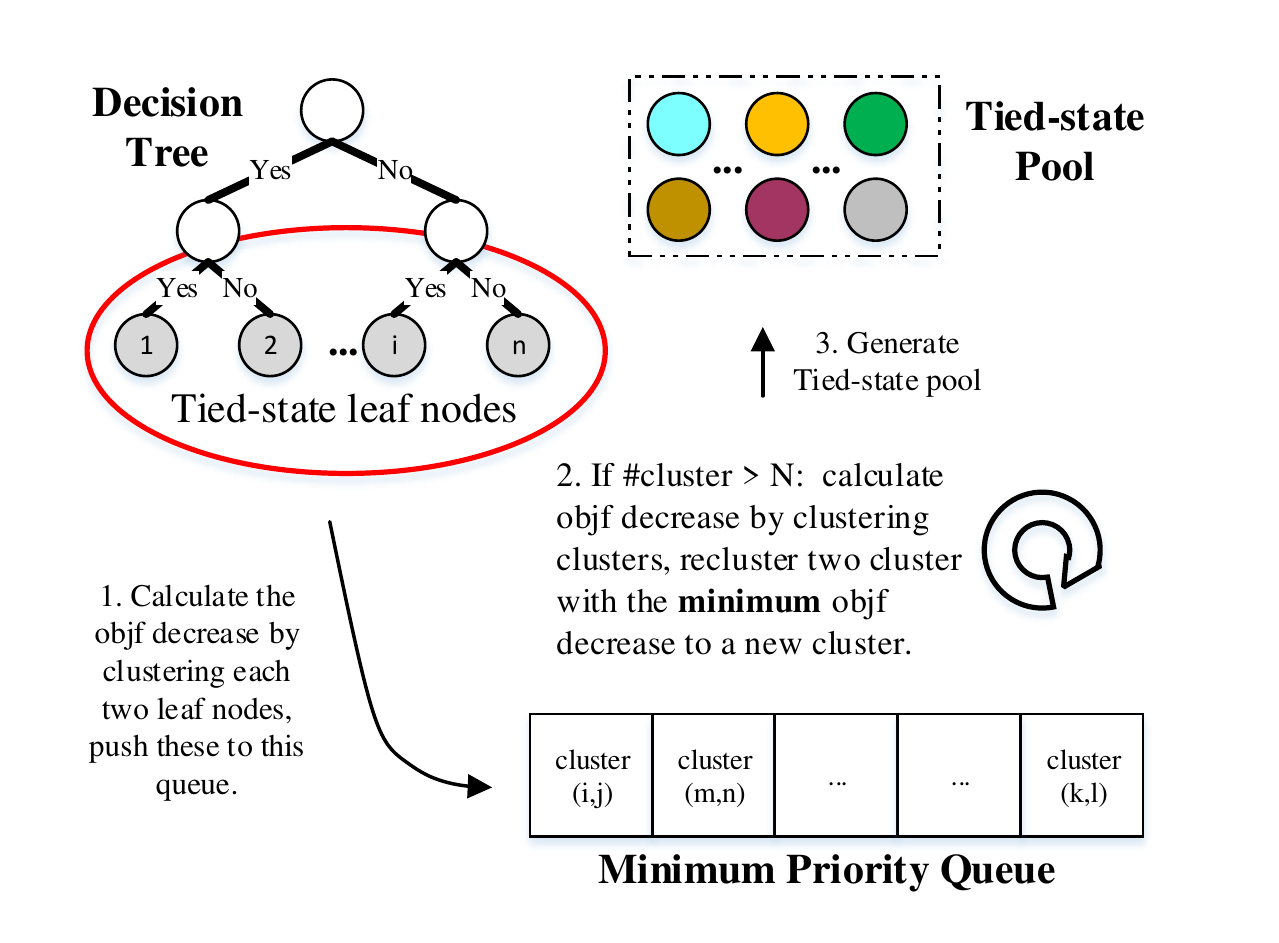}
	\caption{Illustraion of second-step procedure.}
	\label{fig:six}
\end{figure}

In order to get the final tied-state pool, the tied-states generated by first-step are reclusterred in this step. In the second step, the clusters in leaf nodes obtained in the first step is re-clustered by a bottom-up procedure using sequential greedy optimisation. Similar to~\cite{dan2012}, the expected log-likelihood decrease by combining every two clusters is calculated. A minimum priority queue is maintained to re-cluster the two clusters with minimum log-likelihood decrease to a new cluster. This process is repeated until reaching these target tied-state number $\text{N}$. Finally the tied-state pool is composed by the reculusterred tied-state. We illustrate this second-step in Fig.~\ref{fig:six}.

The expected log-likelihood in the above-mentioned two steps can be calculated on the feature vector $\mathbf{x}$ based on the Gaussian distribution assumption withe $D$-dimensional mean vector $\mathbf{\mu}$ and covariance matrix $\mathbf{\Sigma}$:
\begin{eqnarray}
L(\mathbf{x}) &=& E\left[\log \mathcal N\left(\mathbf{x};\mathbf{\mu}, \mathbf{\Sigma}\right)\right]\nonumber\\
~&=&-\frac{1}{2}E\left[(\mathbf{x}-\mathbf{\mu})^{\top}\mathbf{\Sigma}^{-1}(\mathbf{x}-\mathbf{\mu})+\log((2\pi)^D|\mathbf{\Sigma}|)\right]\nonumber\\
~&=& -\frac{1}{2}[(1+\log(2\pi))D + \log|\mathbf{\Sigma}|]
\end{eqnarray}
Let $S$ be a cluster with $N$ training feature vectors, the expected log likelihood on this cluster is given by:
\begin{eqnarray}
L(S) &=& -\frac{N}{2}[(1+\log(2\pi))D + \log|\mathbf{\Sigma}|]
\end{eqnarray}
If we partition $S$ into two subsets $S_1$ and $S_2$, with $N_1$ and $N_2$ feature vectors, mean vectors $\mathbf{\mu}_1$ and $\mathbf{\mu}_2$, covariance matrices $\mathbf{\Sigma}_1$ and $\mathbf{\Sigma}_2$ respectively, then the expected log-likelihood increase after splitting becomes:
\begin{eqnarray}
\Delta L &=& L(S_1)+L(S_2)-L(S) \nonumber \\
&=& \frac{N}{2} \log|\mathbf{\Sigma}| - \frac{N_1}{2} \log|\mathbf{\Sigma}_1| - \frac{N_2}{2} \log|\mathbf{\Sigma}_2|
\end{eqnarray}
Similarly, we can also obtain the expected log-likelihood decrease for the second step re-clustering accordingly. The statistics required in these equations can be calculated from the training data.

\subsection{Data-driven question set generation}
The question set used for state tying is built via a top-down tree-based method like in~\cite{dan2012}. Initially, all characters are placed in root node and the expected log likelihood of all the training data is calculated. Then $k$-means clustering~\cite{hartigan1979algorithm} with $k=2$ is conducted for several times on different initial assignments and the best clustering result is selected to split the root node. A maximum priority queue is maintained to store the likelihood increase by splitting each parent node to two children nodes. This splitting process is recursively performed until each leaf node only has one character class. Each node corresponds to one question which is constituted of all the leaves which this node can reach to when traversing the tree. Finally, the question set consists of these questions.

\section{Experiments}
In this section, we present experiments on recognizing offline handwritten Chinese text line with Kaldi toolkit~\cite{povey2011kaldi}, for the purpose of evaluating and comparing the proposed parsimonious HMMs with the conventional HMMs~\cite{du2016deep}. We use the public CASIA-HWDB database~\cite{liu2011casia} for training, including HWDB1.0, HWDB1.1, HWDB2.0, HWDB2.1, and HWDB2.2 datasets. HWDB1.0 and HWDB1.1 are offline isolated handwritten Chinese character datasets while HWDB2.0-HWDB2.2 are offline handwritten Chinese text datasets. In total, there are 3,980 classes (Chinese characters, symbols, garbage) with 4,091,599 samples. Here ``garbage'' classes represent the short blank model between characters and the long blank model at the beginning or end of the text line. The ICDAR-2013 competition set~\cite{yin2013icdar} is adopted as the evaluation set. The gradient-based feature extracted from one frame of the text line is a 256-dimensional vector, followed by PCA to obtain a 50-dimensional feature vector. This feature vector
is directly used for GMM-HMM systems while an augmented version of 7 frames is fed to DNN-HMM systems.

For GMM-HMM systems, each character class is modeled by a left-to-right HMM with each state modeled by a GMM with 40 Gaussian mixtures. For DNN-HMM systems, the input size of DNN is 350. %The arcithecture of DNN is shown in Fig.~\ref{fig:fuck}. 
The mini-batch size is 256. The initial step size is set to 0.008 which is halved after each iteration if the loss of cross-validation set is reduced. 16 iterations are conducted. 

As for language modeling, 3-gram is adopted and trained with the transcriptions of both the CASIA database and other corpora including 208MB texts of Guangming Daily between 1994 and 1998, 115MB texts of People¡¯s Daily between 2000 and 2004, 129MB texts of other newspapers, and 93MB texts of Sina News. The evaluation measure is CER, which is the ratio between the total number of substitution/insertion/deletion errors and the total number of character samples in the evaluation set. 

%\begin{figure}
%	\centering
%	\setlength{\abovecaptionskip}{0pt}
%	\setlength{\belowcaptionskip}{0pt}
%	\includegraphics[width=0.5\textwidth]{./picture/DNN.pdf}
%	\caption{ DNN structure.}
%	\label{fig:fuck}
%\end{figure}

\subsection{Experiments on different settings of tied-states}

In this subsection, five conventional GMM-HMM systems are built with the fixed number of HMM states per character from 1 to 5. Four parsimonious GMM-HMM (denoted as GMM-PHMM) systems are generated based on the state tying from 5-state GMM-HMM system, yielding average tied-state per character from 1 to 4. Accordingly, five conventional DNN-HMM systems are trained from five conventional GMM-HMM systems while four parsimonious DNN-HMM (denoted as DNN-PHMM) systems are trained based on four GMM-PHMM systems. For DNN-HMM and DNN-PHMM, 6 hidden layers with 2048 nodes for each hidden layer are used and the number of neurons of DNN output layer corresponding to the total number of states varies from 3980 (1 tied-state per character) to 19900 (5 tied-states per character).

\begin{table}[t]
	\begin{center}
		\caption{The CER(\%) comparison of HMM systems with different number settings of tied-states per character $N_s$. } \label{tab:one}
		\newcommand{\tabincell}[2]{\begin{tabular}{@{}#1@{}}#2\end{tabular}}
		\begin{tabular}{|c|c|c|c|c|c|}
			\hline
			$N_s$    & 5 & 4 & 3 & 2 & 1 \\
			\hline GMM-HMM & 20.04 & \textbf{19.94} & 21.94 & 24.92 & 30.34\\
			\hline GMM-PHMM & - &  19.41 & 18.83 & \textbf{18.14} & 18.49\\
			\hline DNN-HMM & \textbf{6.73} & 6.80 & 7.11 & 8.21 & 11.09\\
			\hline DNN-PHMM & - & 6.37 & \textbf{6.31} & 6.48 & 7.15 \\
			\hline
		\end{tabular}
	\end{center}
\end{table}

Table~\ref{tab:one} listed a CER comparison of HMM systems on the evaluation set with different number settings of tied-states per character. Several observations could be made. First, for both GMM-PHMM and DNN-PHMM with the decreasing of the number of tied-states, the CERs first decreased and then increased. This implied that too many states led to the confusion increasing while too few states decreased the discrimination among characters classes. Second, GMM-PHMM/DNN-PHMM systems consistently and significantly outperformed the corresponding GMM-HMM/DNN-HMM systems with the same tied-state number, demonstrating the effectiveness of the proposed state tying algorithm. For example, for the most compact case, namely 1 tied-state per character, GMM-PHMM yielded a relative CER reduction of 39.1\% over GMM-HMM while DNN-PHMM achieved a relative CER reduction of 35.5\% over DNN-HMM. This indicated that the tied-state allocation for different character classes could be much more reasonable after state tying by fully utilizing the similarities among different characters. Finally, in the best configured cases, a relative CER reduction of 9.5\% was achieved by GMM-PHMM over GMM-HMM while a relative CER reduction of 6.2\% was achieved by DNN-PHMM over DNN-HMM. Moreover, 40\% reduction of tied-states in total were obtained in DNN-PHMM compared with DNN-HMM.

\begin{table}[t]
	\begin{center}
		\caption{The CER(\%) comparison of HMM systems with different number settings of tied-states per character $N_s<$1.} \label{tab:two}
		\newcommand{\tabincell}[2]{\begin{tabular}{@{}#1@{}}#2\end{tabular}}
		\begin{tabular}{|c|c|c|c|c|c|}
			\hline
			\tabincell{c}{$N_s$}  & \tabincell{c}{0.9} & \tabincell{c}{0.8} & \tabincell{c}{0.7} & \tabincell{c}{0.6} & \tabincell{c}{0.5}
			\\
			\hline
			\tabincell{c}{GMM-PHMM} & 18.66 & 19.17 & 19.92 & 21.28 & 22.54 \\
			\hline
			\tabincell{c}{DNN-PHMM} & 7.34 & 7.50 & 7.97 & 8.80 & 9.52\\
			\hline
		\end{tabular}
	\end{center}
\end{table}

One more advantage of DNN-PHMM is that we can achieve much more compact design by setting the number of tied-states per character below 1, as shown in Table~\ref{tab:two}. However, for DNN-HMM, the minimum setting is 1 state per character. We could observe from Table~\ref{tab:two} that even in such extreme settings, the recognition performance of GMM-PHMM and DNN-PHMM was gradually declined, not like the sharp decreasing of performance in GMM-HMM and DNN-HMM from 2-state setting to 1-state setting from Table~\ref{tab:one}. With an average 0.5 tied-state per character setting, the corresponding DNN-PHMM outperformed DNN-HMM with 1-state setting and MDLSTM-RNN (with a CER of 10.6\% in \cite{messina2015segmentation}), yielding the relative CER reductions of 14.2\% and 10.2\%, respectively.

\subsection{Experiments on parsimonious modeling}

\begin{table}[t]
	\begin{center}
		\caption{The performance comparison of the best configured DNN-HMM and DNN-PHMM systems with different DNN structures. ($N_{\textrm{U}}$ and $N_{\textrm{L}}$ are the numbers of hidden units and layers, $N_{\textrm{M}}$ and $N_{\textrm{T}}$ are the model size and run-time latency normalized by DNN-HMM with $N_{\textrm{U}}$=2048 and $N_{\textrm{L}}$=6.)} \label{tab:three}
		\newcommand{\tabincell}[2]{\begin{tabular}{@{}#1@{}}#2\end{tabular}}
		\begin{tabular}{|c|c|c|c|c|}
			\hline
			\multicolumn{2}{|c|}{($N_{\textrm{U}}$, $N_{\textrm{L}}$)} &(1024, 4) & (1024, 6) & (2048, 6)
			\\
			\hline
			\multirow{3} * {\tabincell{c}{DNN-HMM}} & CER & 7.15 & 6.91 & 6.73 \\
			\cline{2-5}
			~ & $N_{\textrm{M}}$ & 0.38 & 0.42 & 1\\
			\cline{2-5}
			~ & $N_{\textrm{T}}$ & 0.82 & 0.93 & 1\\
			\hline
			\multirow{3} * {\tabincell{c}{DNN-PHMM}} & CER & 6.78 & 6.48 & 6.31  \\
			\cline{2-5}
			~ & $N_{\textrm{M}}$ & 0.25 & 0.27 & 0.74 \\
			\cline{2-5}
			~ & $N_{\textrm{T}}$ & 0.28 & 0.31 & 0.40 \\
			\hline
		\end{tabular}
	\end{center}
\end{table}

In order to further address the practical issues such as the demand of memory and computation, the performance comparison of the best configured DNN-HMM and DNN-PHMM systems with different DNN structures is listed in Table~\ref{tab:three}. Obviously, with the decrease of hidden units and layers, DNN-PHMM could still maintain a competitive performance while the corresponding model size and run-time latency could be largely reduced. For example, DNN-PHMM using (1024, 4) setting achieved a comparable CER with DNN-HMM using (2048, 6) setting. However, 75\% of model size and 72\% of run-time latency were reduced in DNN-PHMM compared with DNN-HMM. 

\subsection{Results analysis}

\begin{figure}
	\centering
	\setlength{\abovecaptionskip}{0pt}
	\setlength{\belowcaptionskip}{0pt}
	\includegraphics[width=0.5\textwidth]{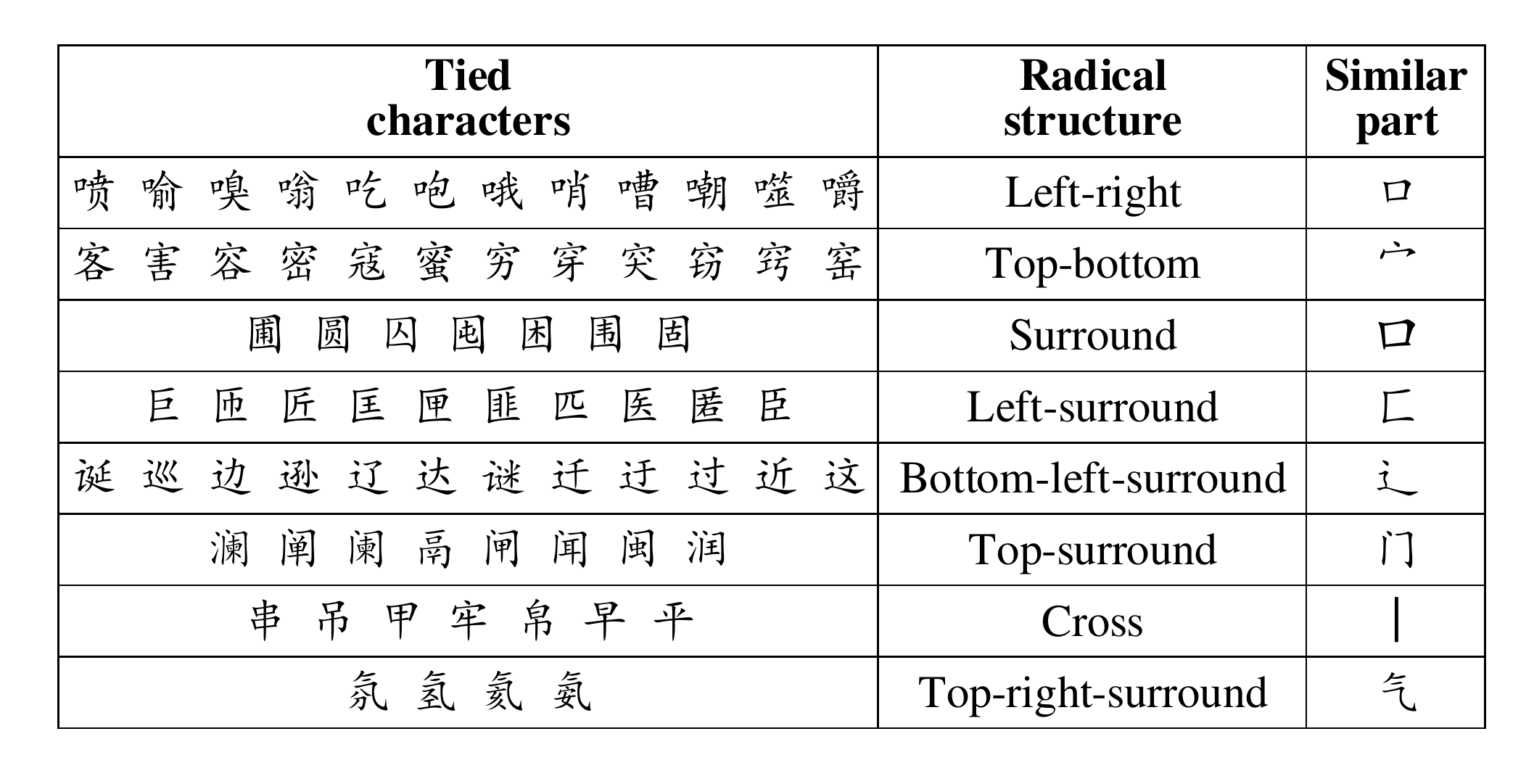}
	\caption{The examples of tied Chinese characters with different radicals and spatial structures.}
	\label{fig:four}
\end{figure}

To explain the reason why the proposed parsimonious HMMs are so effective in parsimonious modeling, we first show the examples of state-tying results in Fig.~\ref{fig:four}. The first column shows the set of tied characters by the state-tying from the first state to the fifth state of 5-state HMM with different radicals structures and similarities described in second and third columns. From these results, we observed that although the vocabulary of Chinese characters could be quite large (tens of thousands), most of them consisted of basic radicals and spatial structures with only a few hundred categories. Accordingly, the Chinese characters with the same or similar radicals were easily tied using the proposed algorithm. This is the reason that the proposed DNN-PHMM with quite compact design can still maintain high recognition performance as shown in Table~\ref{tab:two} and~\ref{tab:three}.

\begin{figure}
	\centering
	\setlength{\abovecaptionskip}{0pt}
	\setlength{\belowcaptionskip}{0pt}
	\includegraphics[width=0.5\textwidth]{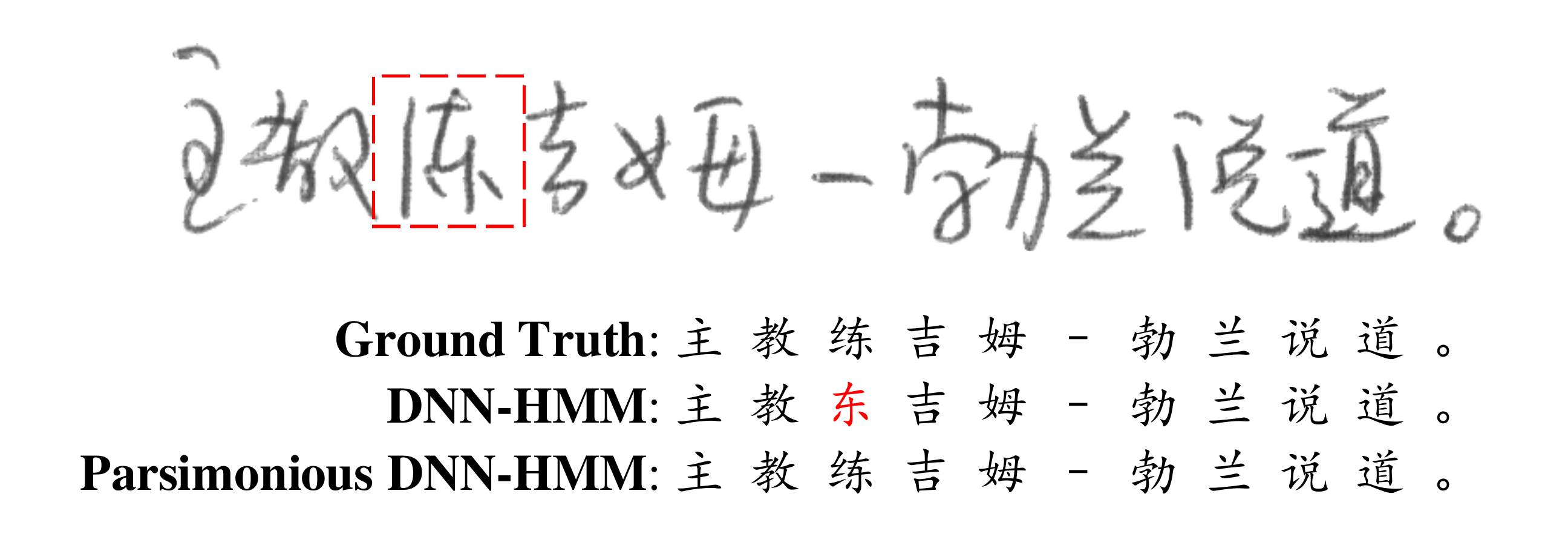}
	\caption{The recognition results comparison between DNN-HMM and DNN-PHMM.}
	\label{fig:five}
\end{figure}

To give readers a better understanding why DNN-PHMM could improve the recognition accuracy over DNN-HMM, a recognition example is shown in Fig.~\ref{fig:five}, where DNN-HMM generates one substitution error (marked red) while DNN-PHMM generates the correct results as the ground truth. This can be explained as: in DNN-HMM system, there are too fewer training handwritten samples with the left radical like the misclassified one in the red dashed box. However, in DNN-PHMM, by state-tying, this unusual writing style of the left radical can be shared from other handwritten Chinese characters samples to train this specific character class.

\section{Conclusion}
In this paper, we present parsimonious DNN-HMMs to reduce model redundancy and capture similarities among different Chinese characters. %Compared with the conventional DNN-HMMs, the proposed parsimonious DNN-HMMs can achieve both more compact design and better recognition accuracy due to the data sharing for the tied states and the confusion decreasing among state classes. 
Note that the model is left-to-right HMM and the features are extracted from left-to-right, so the similarities captured by state tying are more on left-to-right structure. In the future, we plan to investigate the parsimonious modeling for 2D-HMM based HCTR to capture more structure information.


\section*{Acknowledgment}
This work was supported in part by the National Key R\&D Program of China under contract No. 2017YFB1002202, the National Natural Science Foundation of China under Grants No. 61671422 and U1613211, the Key Science and Technology Project of Anhui Province under Grant No. 17030901005, and MOE-Microsoft Key Laboratory of USTC. This work was also funded by Huawei Noah's Ark Lab. The authors would like to thank Mr. Yannan Wang for the contributions on the detail discussion of GMM-HMM.



%\bibliographystyle{IEEEtran}

\end{document}